%% file: main.tex
\documentclass[runningheads]{llncs}
\usepackage[T1]{fontenc}
\usepackage{xcolor}
\usepackage{tikz}
\usepackage{amsmath}
\usepackage{amsfonts}
\usepackage{booktabs,tabularx,multirow,xcolor,colortbl}
\usepackage{graphicx}
\usepackage{makecell}
\usepackage{booktabs,tabularx,multirow,xcolor,colortbl,array,graphicx}
\usepackage{multirow}
\usepackage{subcaption}   
\usepackage{caption}
\usepackage[colorlinks=true, linkcolor=blue, citecolor=green, urlcolor=magenta]{hyperref}

\newcolumntype{Y}{>{\centering\arraybackslash}X}

\newcommand{\paperName}{UniLipi}
\newcommand{\scriptcount}{13}

\usepackage{xcolor}

\definecolor{myBlue}{HTML}{DAE8FC}
\definecolor{myGreen}{HTML}{E8FFE7}
\definecolor{myPurple}{HTML}{E1D5E7}
\definecolor{myOrange}{HTML}{FAD7AC}
\definecolor{myRed}{HTML}{FFCCCC}

\definecolor{borderRed}{HTML}{EA6B66}
\definecolor{borderBlue}{HTML}{6C8EBF}
\definecolor{borderGreen}{HTML}{82B366}
\definecolor{borderPurple}{HTML}{9673A6}
\definecolor{borderOrange}{HTML}{B46504}

\definecolor{imageTokenText}{HTML}{B46504}

\definecolor{scriptTokenText}{HTML}{2E7D32}

\definecolor{countTokenText}{HTML}{6A1B9A}

\newcommand{\circleddashed}[2][]{%
  \tikz[baseline=(char.base)]{
    \node[densely dashed, thick,
      shape=circle,
      draw=black, fill=white, text=black,
      font=\sffamily,
      inner sep=1pt,
      minimum size=2pt,
      #1] (char) {#2};
  }
}

\definecolor{northcentralcolor}{HTML}{4A86E8}   
\definecolor{southcolor}{HTML}{E06666}          
\definecolor{himalayancolor}{HTML}{8E7CC3}      
\definecolor{westerncolor}{HTML}{6AA84F}        
\definecolor{easterncolor}{HTML}{F6B26B}        

\begin{document}
\title{UniLipi: A Unified Multi-Script OCR for Historical Indic Manuscripts}

\author{
Tathagata Ghosh\orcidID{0009-0009-6183-7393},
Sai Madhusudan Gunda\orcidID{0009-0003-7226-3635}, \\
Simran Singh Sandral\orcidID{0009-0003-6256-9286},
Ravi Kiran Sarvadevabhatla\orcidID{0000-0003-4134-1154}\\
\begin{tabular}{c c}
{\tt\small tathagata.ghosh@research.iiit.ac.in,} &
{\tt\small sai.gunda@research.iiit.ac.in,} \\
{\tt\small ssandral@gitam.in,} &
{\tt\small ravi.kiran@iiit.ac.in}
\end{tabular}
}
\authorrunning{Tathagata et al.}
%
\institute{International Institute of Information Technology Hyderabad, INDIA\\
\url{https://ihdia.iiit.ac.in/unilipi/}}
%
\maketitle              

\begin{abstract}
Optical character recognition (OCR) for handwritten Indic manuscripts is essential for large-scale digitization and computational access to manuscript heritage. However, existing approaches are typically developed for one script at a time and require substantial script-specific customization. This limits scalability and practical deployment across diverse collections. We present \paperName, a unified multi-script OCR model for handwritten Indic manuscripts trained jointly across \scriptcount\ Indic scripts within a single framework. \paperName\ directly handles realistic manuscript conditions, including extreme variation in line geometry, large variation in line length, and partial interruptions caused by non-textual manuscript entities such as holes, stains, or pictorial illustrations. To operate effectively under ultra low-resource conditions, the model leverages script-aware synthetic manuscript data generation, substantially reducing reliance on large volumes of real annotated data. Beyond historical manuscripts, we show that \paperName\ serves as an effective foundational pretrained model. Specifically, its learned representations enable good OCR performance for contemporary Indic handwriting and extend to several non-Indic scripts, including Tibetan, Italian, Latin, and Chinese scripts. In addition to transcription, \paperName\ predicts script identity and per-line native character counts, supporting practical manuscript cataloging workflows.

\keywords{Historical OCR \and Handwritten Text Recognition \and Multi script OCR}
\end{abstract}

\section{Introduction}
\label{sec:intro}

Optical character recognition (OCR) for handwritten Indic manuscripts is a critical enabler for large-scale digitization, preservation, and computational access to Indic documentary heritage, which spans centuries, regions, and institutional contexts. Beyond individual archives, coordinated efforts increasingly seek OCR systems that can operate robustly across diverse collections, scripts, and material conditions. In this context, mission-mode national initiatives such as GyanBharatam~\cite{gyanbharatam_website} articulate the need for systematic manuscript digitization, and open challenges such as GyanSetu~\cite{gbm_gyansetu_website} have explicitly explored how AI can be brought into service to meet these requirements. Despite recent progress in handwritten text recognition, OCR for Indic manuscripts remains fragmented, script-specific, and difficult to deploy at scale.

\begin{figure}[!t]
    \centering
    \includegraphics[width=\textwidth]{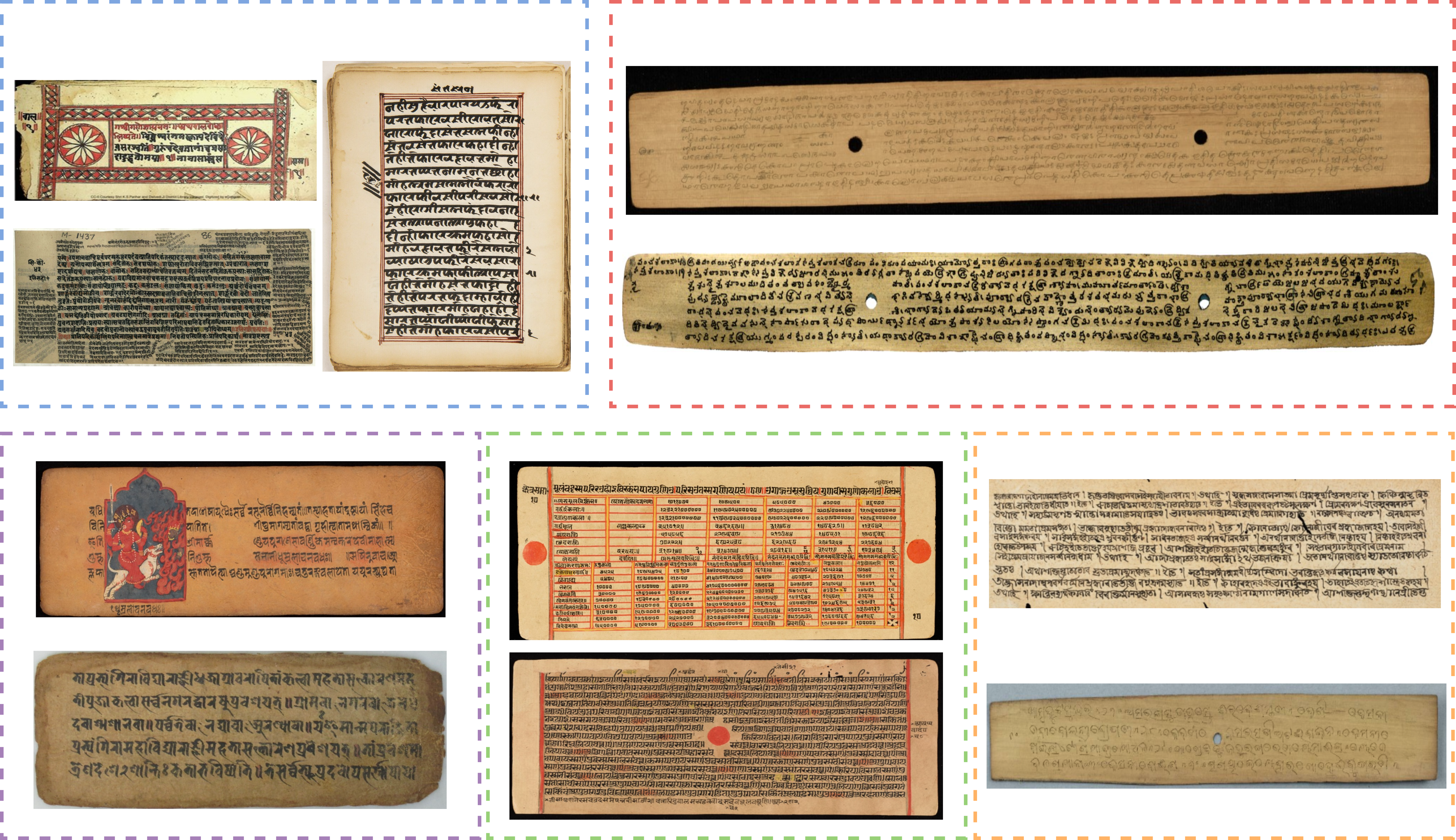}
\caption{\textbf{Diversity of manuscript styles across Indic script regions.}
Examples of manuscript lines from multiple historical collections grouped by geographic and script traditions:
\textcolor{northcentralcolor}{North/Central Indian} ,
\textcolor{southcolor}{Southern Indian} , 
\textcolor{himalayancolor}{Himalayan/Buddhist} ,
\textcolor{westerncolor}{Western Indian} ,
 and
 \textcolor{easterncolor}{Eastern Indian}.
The examples illustrate substantial diversity in writing substrates (palm leaf, paper), layout geometry, stroke thickness, ink characteristics, and orthographic structure across scripts and regions.
This diversity motivates \paperName\ model and reference-conditioned synthesis pipeline, which extracts appearance and geometric priors from real manuscripts to generate realistic synthetic training data.}

    \label{fig:diversemanuscirpt_images}
\end{figure}

A fundamental challenge is the \emph{ultra low-resource} nature of historical handwritten data. Many Indic scripts are represented by a small number of manuscript collections, often with limited annotations, high intra and inter-writer variability, and substantial variation in manuscript condition~\cite{Prusty2019,phd11}. In addition, the scripts and collections pose difficult recognition challenges, including dense ligatures and diacritics, visually similar glyph inventories across scripts, and diverse writing materials and degradation patterns (see Fig. \ref{fig:diversemanuscirpt_images}). As a result, most existing approaches focus on a \emph{single} script or collection at a time, carefully tailoring model architectures, training procedures, and post-processing steps to a specific script~\cite{sharma-etal-2025-ancidev,Kausadikar2025MoDeTrans,Buoy2023KhmerBaseline,JindalGhosh2021,Gunda_2026_WACV}. This one-script-at-a-time paradigm imposes significant practical limitations. From a model development perspective, it requires maintaining separate OCR systems for each script, including independent training runs, hyperparameter tuning, and evaluation pipelines. From a deployment and user perspective, it leads to brittle workflows in which OCR capability does not generalize across collections, even when manuscripts share similar linguistic or structural properties. Consequently, large-scale digitization efforts continue to rely on a growing set of script-specific OCR systems, limiting scalability and long-term maintainability.

An important observation is that many Indic manuscripts often encode the \emph{same} underlying language, most prominently Sanskrit~\cite{namami_scripts_languages}. While scripts differ in visual form, the target language and orthographic structure are often shared. This suggests that learning to recognize handwritten Indic text should not necessarily be framed as a collection of isolated script-specific problems, but rather as a \emph{unified} learning problem with shared representations across scripts.

Inspired by this observation, we introduce \textbf{\paperName}, a unified multi-script OCR model for handwritten Indic manuscripts.  \paperName\ is designed to operate directly on realistic manuscript data, handling complex line geometries including strongly curved text, variable line lengths, and interruptions caused by holes, stains, or embedded illustrations. To address the ultra low-resource nature of many historical scripts, we train the model using a combination of limited annotated real data and script-aware synthetic manuscript lines that mimic authentic writing conditions. The model is jointly trained across multiple Indic scripts and manuscript collections, enabling it to learn shared visual representations while remaining sensitive to script-specific characteristics. By leveraging both cross-script similarities and structural diversity, \paperName\ generalizes beyond peculiarities of individual scripts, allowing a single OCR system to operate across a broad range of historical manuscripts.

We also show that \paperName\ serves as an effective foundational pretrained model beyond historical Indic manuscripts. Specifically, \paperName\ enables OCR for contemporary Indic handwriting~\cite{phd11} and extends to several non-Indic scripts.

\vspace{6pt}
\noindent\textbf{Our main contributions are:}
\vspace{-6pt}
\begin{itemize}
\item \paperName, the first unified foundational multi-script OCR model for handwritten Indic scripts, trained jointly across \scriptcount\ distinct Indic scripts within a single framework.
\item A script-aware synthetic manuscript data generation approach that reduces reliance on large volumes of real annotated handwritten data.
\item An OCR system that, in addition to transcription, predicts script identity and per-line glyph counts.
\end{itemize}

\section{Related Work}
\label{sec:related}

\noindent\textbf{Historical OCR/HTR:} Modern handwritten text recognition (HTR) has largely shifted to sequence modeling, including CNN-RNN, LSTM, and Transformer-based recognizers~\cite{CRNN,Li2021TrOCR,Dwivedi2020}. While autoregressive encoder--decoder models have gained popularity in recent OCR systems \cite{10.1007/978-3-032-04614-7_7,ChewUnified2025}, CTC-based formulations remain competitive for long and irregular text lines, particularly in low-resource settings~\cite{garrido2025generalization}. Many HTR systems are evaluated on relatively homogeneous corpora or short word-level crops, and rectification modules originally developed for scene text~\cite{Shi2018ASTER,Luo2019MORAN} are typically assessed under limited curvature and cleaner conditions.  These constraints motivate recognition frameworks that explicitly account for writing background variability and long-line geometry.

\noindent\textbf{Low-resource and Indic manuscript OCR:} Prior work has explored transfer learning, synthetic pretraining, and cross-lingual representation sharing to mitigate data scarcity in both printed and handwritten settings~\cite{baek2021wrong,rossetta2019,al2019arabic,Souibgui2022FewShots}. However, certain historical manuscripts introduce additional domain gaps due to background variability, degradation, and long, contiguous and irregular text lines~\cite{Prusty2019,Valy2018SleukRith}, limiting the effectiveness of models pretrained on modern corpora. Historical Indic manuscripts frequently contain contiguous text with limited spacing, dense ligatures and diacritics, and highly variable line lengths, making segmentation-free full-line recognition substantially more challenging. Recent efforts in Indic and South-east Asian manuscript OCR demonstrate the feasibility of deep models in data-scarce settings~\cite{sharma-etal-2025-ancidev,Kausadikar2025MoDeTrans,Chandure2021Modi,JindalGhosh2021,Gunda_2026_WACV}. However, script-specific training and output handling remain common practice, requiring separate models for each script. This motivates unified strategies that reduce per-script overhead while remaining robust to manuscript-specific variability.

\noindent\textbf{Synthetic data for manuscript OCR:}
Synthetic corpora are widely used to mitigate annotation scarcity in OCR and handwritten text recognition (HTR). A recent survey categorizes handwriting synthesis approaches into reference based line generation and text to handwriting generation models~\cite{Diaz2025Survey}. Early strategies construct synthetic lines from isolated characters or annotated reference layouts~\cite{Shen2016}, while deformation-based augmentation methods perturb existing samples using geometric and photometric transformations~\cite{Wigington2017,Kang2020}. More recent generative approaches leverage adversarial , transformer and diffusion models to synthesize~\cite{Vogtlin2021GAN,Bhunia_2021_ICCV,Pippi2025,Zaccagnino2026,Dai2025,10.1145/3637528.3671589}. But these require training auxiliary networks which demand substantial data and computational resources. In contrast, ultra low-resource historical manuscript settings require controllable and data-efficient synthesis mechanisms that can explicitly incorporate background texture, physical damage, and complex line curvature. We therefore adopt a hybrid, reference conditioned yet training free synthesis pipeline for \paperName\ that preserves structural realism while remaining computationally lightweight. 

\noindent\textbf{Multi-script and multilingual text recognition:} Multilingual text recognition has been explored in both printed and handwritten settings, often combining script identification with language-specific recognition.  Early multi script OCR systems relied on dedicated script identification modules, using script prediction as a precursor to language specific recognition pipelines~\cite{Genzel2013ScriptID}. Subsequent approaches moved towards proposing shared architectures that jointly perform script identification and multilingual handwriting recognition~\cite{Wang2020MuLTReNet}. More recent methods use incremental and continual learning strategies for adding languages~\cite{Zheng2023MRN,Liu2024HAMMER,Dhiaf2025UCLMHTR}. But these systems primarily use multilingual recognition as routing mechanism with merged vocabularies and language aware decoding. In contrast, the Indic manuscript setting is predominantly multi-script but frequently single-language, with texts such as Sanskrit written across diverse scripts. This distinction motivates unified modeling across scripts within a single decoding space rather than cascaded routing or incremental expansion.

\definecolor{visualtokenbg}{HTML}{FFE6CC}
\definecolor{scripttokenbg}{HTML}{D5E8D4}
\definecolor{counttokenbg}{HTML}{E1D5E7}

\newcommand{\visualtokenscolor}[1]{\colorbox{visualtokenbg}{#1}}
\newcommand{\scripttokencolor}[1]{\colorbox{scripttokenbg}{#1}}
\newcommand{\counttokenscolor}[1]{\colorbox{counttokenbg}{#1}}

\begin{figure}[!t]
    \centering
    \includegraphics[width=\textwidth]{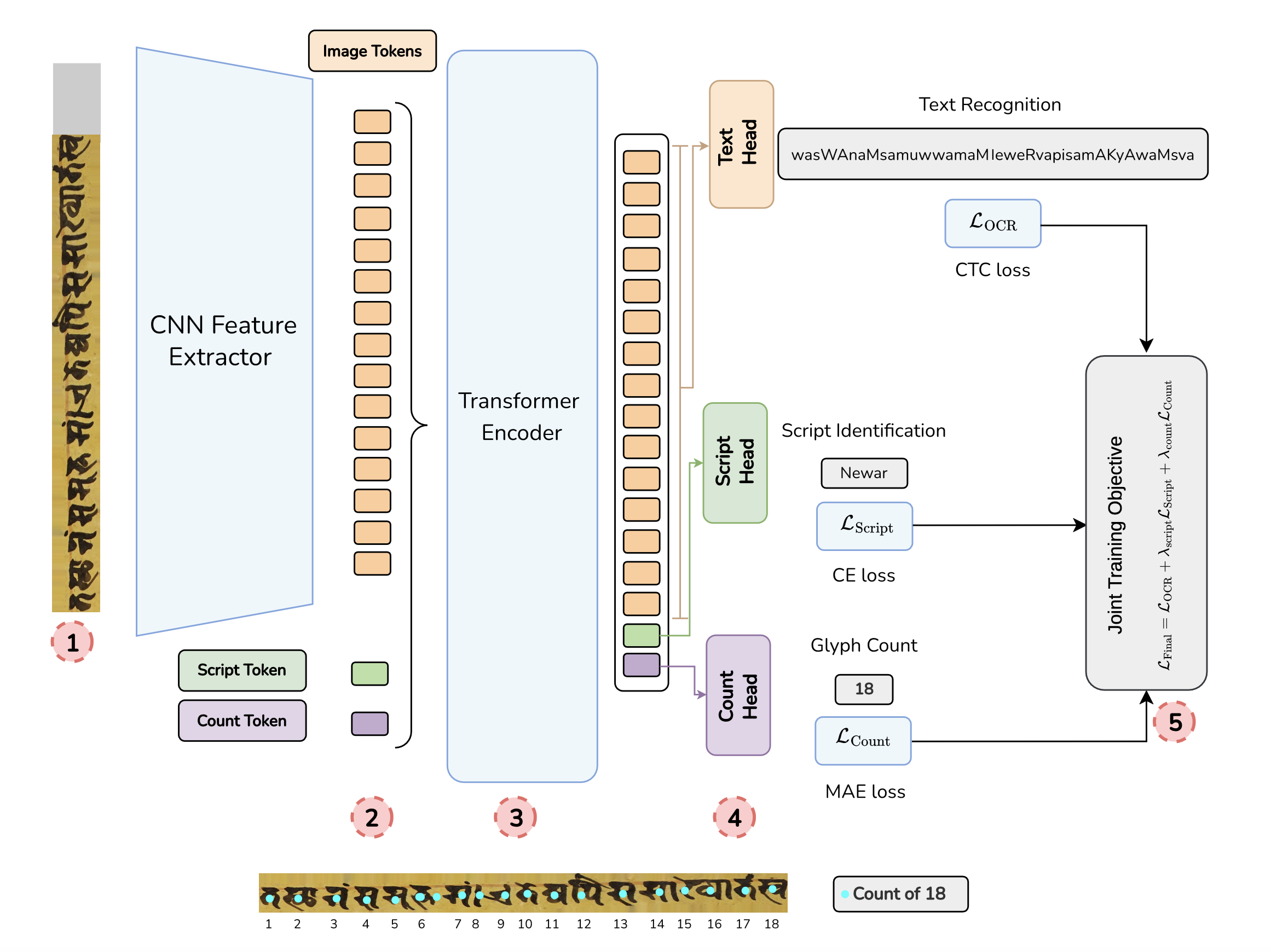}
\caption{\textbf{\paperName\ architecture overview.}
Given a padded line image \circleddashed[draw=borderRed, fill=myRed , text=black]{1}, 
a CNN backbone extracts a sequence of \visualtokenscolor{image tokens}. 
A learnable \scripttokencolor{script token} and a 
\counttokenscolor{count token} are appended to \visualtokenscolor{image token} sequence. The combined token sequence \circleddashed[draw=borderRed, fill=myRed , text=black]{2} is encoded with a Transformer encoder \circleddashed[draw=borderRed, fill=myRed , text=black]{3}. Finally, the image-token representations are decoded by a CTC-based \visualtokenscolor{text head} to produce a Roman \texttt{WX} transcription. The latent task token outputs are fed to respective heads for \scripttokencolor{script identification} and \counttokenscolor{glyph count prediction} \circleddashed[draw=borderRed, fill=myRed , text=black]{4}. The model is trained with a joint objective combining ocr, script, and count losses \circleddashed[draw=borderRed, fill=myRed , text=black]{5}.}

    \label{fig:modelfigure}
\end{figure}

\section{Methodology}
\label{sec:method}

\textbf{Overview:} \paperName\ is a unified multi-task OCR model which recognizes historical manuscript lines exhibiting complex visual structure, multi-script variability, and highly non-uniform character spacing. Given a manuscript line image $I \in \mathbb{R}^{H \times W \times 3}$, the model jointly predicts three complementary outputs:

\begin{itemize}
    \item a text sequence, expressed in the unified Roman \texttt{WX} representation \footnote{ transliteration scheme for representing Indian languages} \cite{Gupta2010TransliterationAI}

    \item a script identity label $s \in \{1, \dots, K\}$ 

    \item a glyph count $c \in [0, N_{\max}]$ corresponding to the number of visually distinct written units in the line.
\end{itemize}

Roman \texttt{WX} character counts do not reflect visual structure in Indic scripts due to conjunct formation and diacritic attachment. Therefore, we define $c$ as the number of glyphs visually distinct graphical units present in the line. This auxiliary objective encourages length aware and spatially precise representations, improving alignment and text recognition for long manuscript lines (Sec \ref{sec:ablations}). All three tasks are learned jointly within a unified architecture, enabling efficient parameter sharing and consistent representation learning across scripts (see Fig \ref{fig:modelfigure}).

\textbf{Architecture:} Our architecture adopts a hybrid CNN-Transformer with task specific heads.
\noindent\textit{Visual Feature Extraction:} Input images are resized to a fixed height $H=68$ and padded to a maximum width $W_{\max}=3200$, for all line lengths. We employ a ResNet~\cite{He_2016_CVPR} backbone to extract visual features ( see Fig \ref{fig:modelfigure} \circleddashed[draw=borderRed, fill=myRed , text=black]{1} ). To transform the 2D feature map $X \in \mathbb{R}^{h \times w \times d}$ into a 1D sequence suitable for the Transformer, we perform vertical pooling followed by a learned linear projection
resulting in a sequence of $T$ image tokens $[{e}_1, \dots, {e}_T]$. \textit{Multi-Task Transformer Encoder:} 
To enable auxiliary supervision without increasing overhead, we introduce two learnable latent task tokens: ${e}_{\text{script}}$ and ${e}_{\text{count}}$. These are concatenated to the image tokens before sending to the Transformer encoder. The input to the transformer is  defined as 
$[{e}_{\text{script}}, {e}_{\text{count}}, {e}_1, \dots, {e}_T]$ \circleddashed[draw=borderRed, fill=myRed , text=black]{2}.
As these tokens participate in the self attention mechanism in the transformer  \circleddashed[draw=borderRed, fill=myRed , text=black]{3} , they aggregate global contextual information from the image tokens. Finally the transformer outputs the following representations $[{h}_{\text{script}}, {h}_{\text{count}}, {h}_1, \dots, {h}_T]$. To efficiently process long sequences, we use Windowed Self-Attention , where each token attends only to neighboring tokens within a fixed window of size $w$ reducing the computational complexity \cite{child2019generatinglongsequencessparse}. Here, transformer encoder is of 6 layers and 8 heads.

We configure the encoder with shared output heads for respective tasks \circleddashed[draw=borderRed, fill=myRed , text=black]{4}. \textit {Text Head:} The OCR text is predicted from $[{h}_1, \dots, {h}_T]$ using a linear projection trained with CTC objective~\cite{connectionist_graves_2006}. At inference time, the transcription is obtained via greedy CTC decoding. \textit{Script Head:} The script identity label  is predicted from ${h}_{\texttt{script}}$ using a $K$-way classifier trained with cross-entropy objective. \textit{Count Head:} The glyph count is predicted from ${h}_{\texttt{count}}$  trained as regression task with MAE objective. 

\paragraph{Training Objective.}
The model is trained end-to-end using a joint optimization objective defined as
\begin{equation} 
\mathcal{L}
=
\mathcal{L}_{\mathrm{OCR}}
+
\lambda_{\mathrm{script}} \mathcal{L}_{\mathrm{script}}
+
\lambda_{\mathrm{count}} \mathcal{L}_{\mathrm{count}}
\end{equation}
where $\mathcal{L}_{\mathrm{OCR}}$ is CTC loss for text recognition, 
$\mathcal{L}_{\mathrm{script}}$ is a cross-entropy loss for script identification, 
and $\mathcal{L}_{\mathrm{count}}$ is a mean absolute error (MAE) loss for glyph count. 
The coefficients $\lambda_{\mathrm{script}}$ and $\lambda_{\mathrm{count}}$ are set empirically and are fixed across all experiments. This joint training strategy encourages the encoder to learn richer representations that jointly capture textual content, script identity, and structural length information, thereby improving transcription accuracy across diverse manuscripts.

\section{Real Manuscripts Data}
\label{sec:data}

Our manuscript collection prioritizes diversity, drawing from public and private collections \cite{indic_man_archive,namami,DATA/WI9184_2022} spanning palm-leaf folios, handmade paper, and religious archives. To ensure pan-Indic coverage, we curate data across \scriptcount\ scripts from five regional traditions: North/Central Indian (Devanagari, Sharada, Modi), Western Indian (Gurmukhi, Jaini), Eastern Indian (Bengali, Odia), Southern Indian (Grantha, Telugu, Malayalam, Kannada), and Himalayan/Buddhist (Newar, Siddham). Together, these scripts span shirorekha\footnote{\textit{shirorekha} refers to a horizontal line connecting individual letters in some scripts.} and non-shirorekha writing systems, as well as rounded, angular, and cursive glyph forms (see Fig \ref{fig:diversemanuscirpt_images}). Due to ultra-low-resource and rare nature of several scripts, the amount of annotated data varies substantially, ranging from 50 to 600 pages per script. This imbalance arises from the challenge of sourcing reliable annotations.

For the manuscript pages, we run a line segmenter~\cite{Agrawal2024LineTR} to extract line images. For unified training, we convert all text to a unified Roman \texttt{WX} representation using a deterministic, lossless mapping~\cite{AksharamukhaPyPI}. This ensures that every orthographic unit in the original script is preserved without ambiguity while enabling a single shared vocabulary across scripts. The resulting \texttt{WX} sequence serves as the sole transcription target for \paperName\ supervision, allowing the model to perform script-agnostic recognition while retaining full reversibility back to the original script. 
 We also record the script name and number of letter conjuncts present in each line (glyph count) as auxiliary output targets during training.

\begin{figure}[h]
    \centering
    \includegraphics[width=\textwidth]{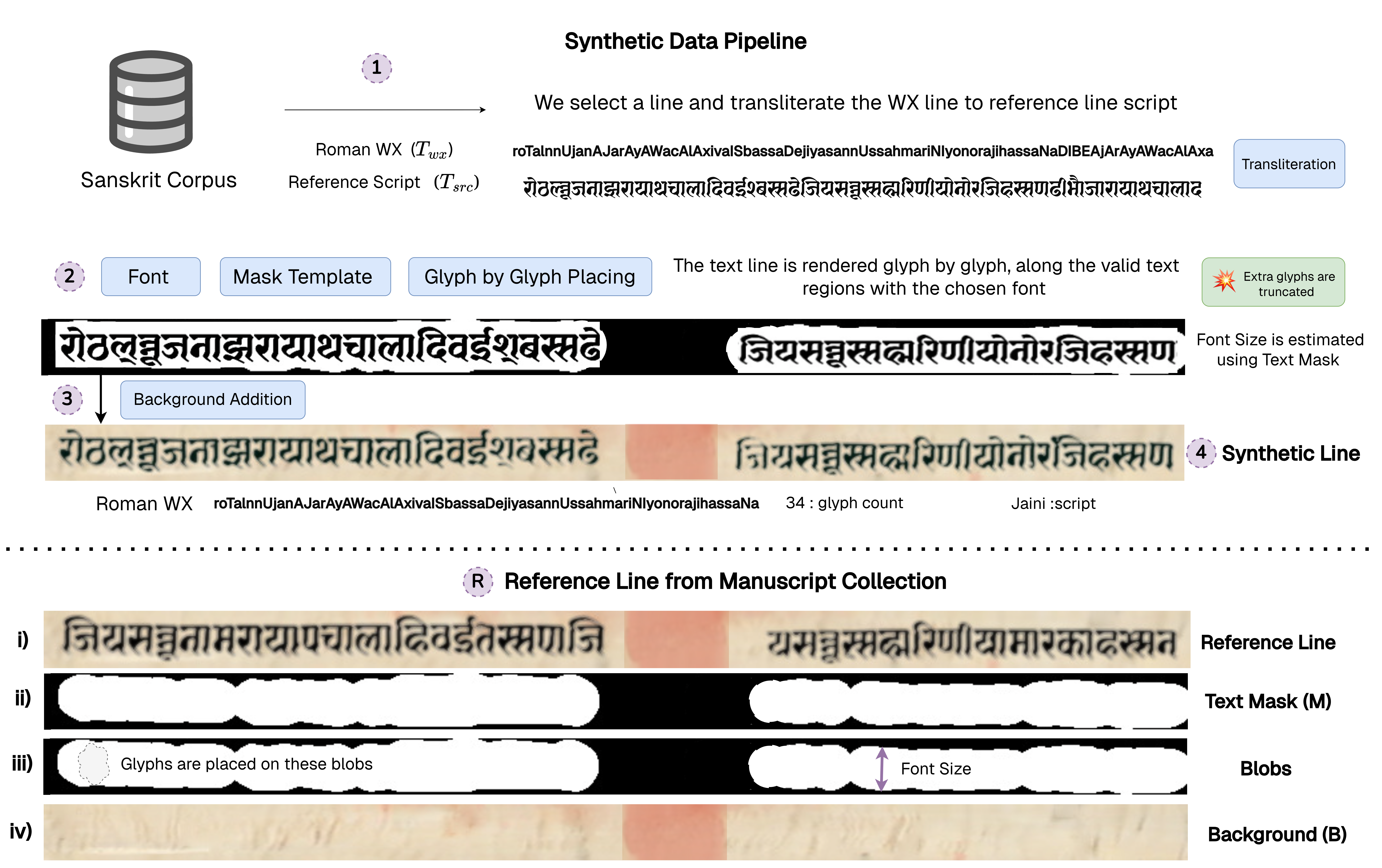}
\caption{\textbf{Reference-conditioned synthetic line generation pipeline.} 
\circleddashed[draw=borderPurple, fill=myPurple, text=black]{1} 
A text line is sampled in the unified Roman \texttt{WX} representation and transliterated into the target script to define the visual content. 
\circleddashed[draw=borderPurple, fill=myPurple, text=black]{2} 
Using a reference manuscript line \circleddashed[draw=borderPurple, fill=myPurple, text=black]{R}, we extract the writing mask, estimate font size, font color , and render glyphs sequentially within valid regions using script specific fonts, while glyphs exceeding are truncated. 
\circleddashed[draw=borderPurple, fill=myPurple, text=black]{3} 
The rendered text is composited onto the clean background to preserve realistic manuscript characteristics. 
\circleddashed[draw=borderPurple, fill=myPurple, text=black]{4} 
The final synthetic line is produced along with its Roman \texttt{WX} transcription target, script identity label, and glyph count supervision.}

    \label{fig:syntheticdiagramfigure}
\end{figure}

\section{Synthetic Manuscripts Data}
\label{sec:synthetic}

The availability of real annotated manuscript lines across all \scriptcount\ scripts is extremely limited. To enable large-scale training while preserving manuscript realism, we develop a pipeline that generates synthetic line images by utilizing  a reference manuscript line image for appearance and geometric priors.

\textit{Text corpus:} We first construct a large text corpus from publicly available ancient Indian textual sources spanning diverse scripts and linguistic traditions~\cite{GRETILSite,DCSHellwig,MangalamCorpus,sanskritdocuments}. All text is converted into the unified Roman \texttt{WX} representation. During synthetic data generation, we sample a text line from this corpus. This line is transliterated to reference manuscript's script for rendering ($T_{\text{scr}}$ - see Fig. \ref{fig:syntheticdiagramfigure} \circleddashed[draw=borderPurple, fill=myPurple, text=black]{1}). 

\textit{Text mask and background preparation:} Our synthesis process begins by sampling a reference manuscript line  image \circleddashed[draw=borderPurple, fill=myPurple, text=black]{R} belonging to a script $s$. We apply a text segmentation model~\cite{ye2025hi} on the line image to identify text regions and obtain a mask $\mathbf{M}$ representing valid writing locations. The original text is then removed using text guided diffusion-based inpainting~\cite{Rombach_2022_CVPR}, producing a clean background image $\textbf{B}$ that preserves background texture and aging patterns of the manuscript.

\textit{Mask-guided text placement:} The writing mask $\mathbf{M}$ is decomposed into writing blobs, each representing a valid text placement region. The transliterated script text $T_{\text{scr}}$ is segmented into glyphs and allocated across blobs based on their spatial capacity. Text is rendered glyph-by-glyph on the valid blobs on a white image with the same dimensions of the reference image using the estimated ink color $C_{\mathrm{ink}}$. Placement is strictly constrained by the mask, ensuring that the ink appears only within valid writing regions (see Fig. \ref{fig:syntheticdiagramfigure}  \circleddashed[draw=borderPurple, fill=myPurple, text=black]{2}). This process naturally determines the number of visible glyphs. Although an initial string of length up to $N_{\max}$ is sampled, glyphs that exceed spatial length are truncated during placement. The final number of rendered glyphs gives us the glyph count. 

\textit{Font color and size estimation:} The font size is automatically determined from blob height, ensuring consistency with the manuscript's original scale. We estimate the ink color distribution directly from the reference image. Foreground stroke pixels are isolated, and channel-wise median RGB values ($C_{\mathrm{ink}}$) are computed. This provides a robust estimate of the ink color, accounting for fading, absorption, and scanning conditions. We render using only the fonts which mimic the ancient style of writing for that reference script.

\textit{Final training sample:} The text line (see Fig. \ref{fig:syntheticdiagramfigure} \circleddashed[draw=borderPurple, fill=myPurple, text=black]{3}) is composited with background image to obtain the final rendered synthetic image. To further reduce the distribution gap between synthetic and real manuscripts, we apply manuscript specific augmentations~\cite{info11020125,augraphy_library}. In summary, each synthesized line image $\mathbf{I}$ (Fig. \ref{fig:syntheticdiagramfigure} \circleddashed[draw=borderPurple, fill=myPurple, text=black]{4}) is paired with (i) Text target in \texttt{WX} representation (truncated) (ii) Script identity label $s \in \{1,\dots,K\}$ (iii) Glyph count $c$. 
By deriving all appearance, geometric, and structural priors from real manuscript references, the resulting synthetic corpus closely matches real manuscripts while enabling scalable training.

\section{Experimental Setup}
\label{sec:experiments}

\textbf{Data:} Using the rendering pipeline described in Sec.~\ref{sec:synthetic}, we generate approximately 2.5M synthetic line images per script, resulting in a total of 32.5M synthetic samples across all \scriptcount\ scripts. A held-out synthetic validation set is used for early stopping during pretraining. Across the \scriptcount\ scripts, the number of available real manuscript line image samples typically varies from $500$ to $3000$. We fine-tune on the training split (90\%) and evaluate on the test split (10\%) of real manuscript data. The splits are done at page level rather than line level to minimize overfitting. 

\noindent\textbf{Training and evaluation setup:} All baselines including \paperName\ are trained using a two-stage procedure. Models are first trained on the synthetic corpus. Early stopping is applied using a held-out synthetic validation split. This pretrained model is then jointly finetuned on the combined real training sets across all scripts, enabling adaptation to real manuscripts. We use the Adam optimizer~\cite{kingma2014adam} with an initial learning rate of $2\times10^{-4}$, learning rate of $10^{-5}$, weight decay of $0.01$ and dropout rate of $0.1$. $\lambda_{\mathrm{count}}$ , $\lambda_{\mathrm{script}}$ are set to 0.4 empirically.  Training is performed using bf16 mixed precision on 4$\times$ NVIDIA H100 GPUs. \paperName\ takes 0.3 seconds for decoding on each line. To assess cross-domain generalization, we additionally evaluate \paperName\ on publicly available text recognition benchmarks.

\noindent \textbf{Evaluation Metrics:} We report Character Error Rate (CER) as the primary ocr  evaluation metric. CER is computed after converting model predictions from Roman \texttt{WX} representation back to native Unicode script. Since manuscript lines typically lack word boundaries, CER serves as the primary metric. Word Error Rate (WER) is additionally reported on benchmarks where word segmentation is available. Both CER and WER are shown on the scale of 0 - 100. We evaluate script classification using macro-averaged F1 score across all script classes, ensuring equal weighting across scripts regardless of dataset size. We evaluate glyph count prediction using mean absolute error (MAE) between predicted and ground-truth glyph counts. All reported overall scores are weighted average based on dataset size.

\noindent \textbf{Baselines:} We compare \paperName\ against representative handwritten text recognition (HTR) systems including LSTM based models  ~\cite{TesseractOCR,kraken2023,CRNN}, and 
Transformer based models \cite{Barrere2024VLT,truc2025htr,Li2021TrOCR,Li2025HTRVT}. All baselines are trained using their official implementations and following same conditions as \paperName.

\input{tables/unilipi15}
\input{tables/baselines}
\input{tables/t1}

\begin{figure}[t]
    \centering
\includegraphics[width=0.85\linewidth]{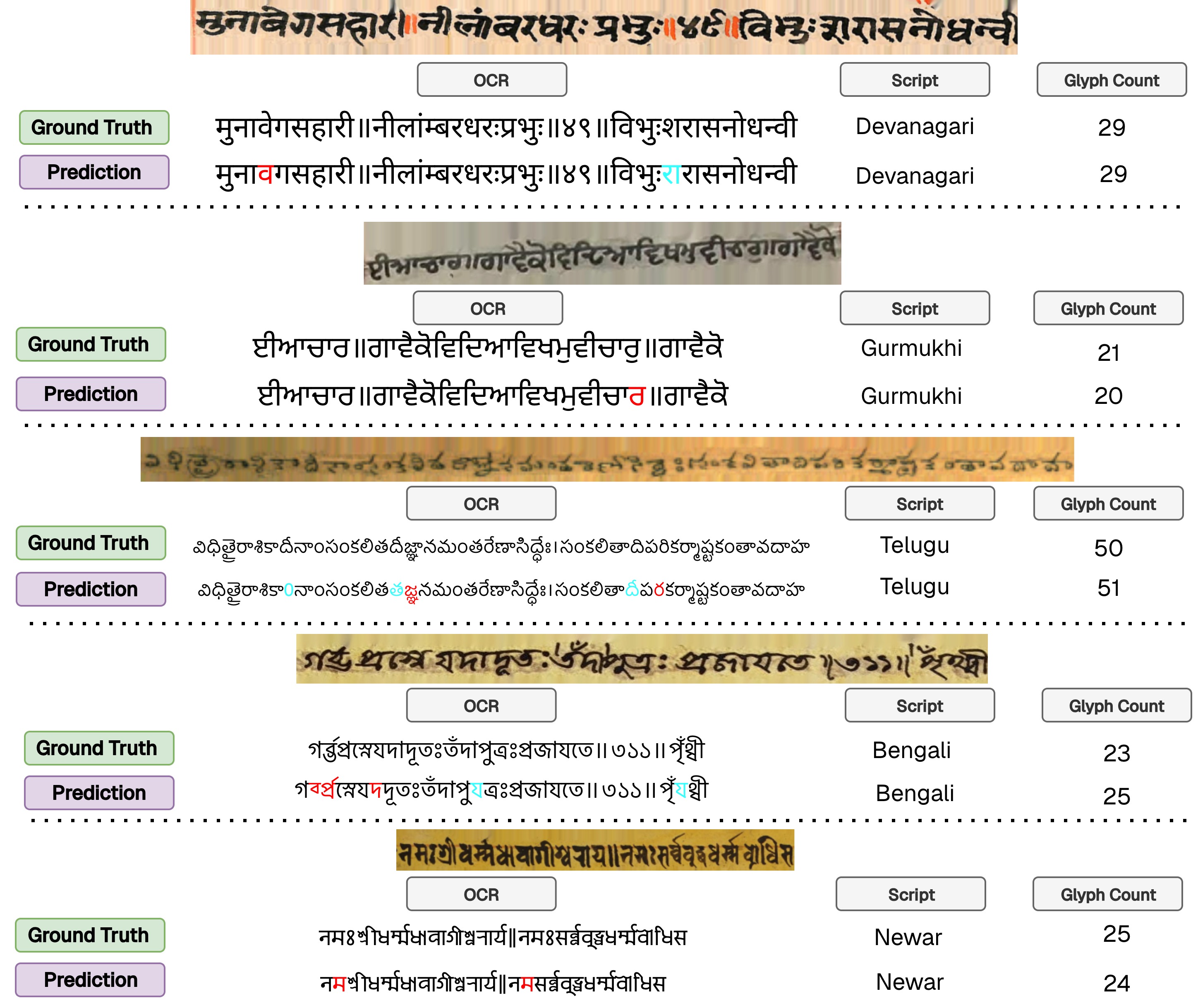}
    \caption{\textbf{Qualitative results} of \paperName{} on script for each region,
    \textcolor{red}{red means deletion},
    \textcolor{cyan}{blue means substitution or addition}.}
    \label{fig:qualitativeresults}
    \vspace{-10pt}
\end{figure}

\input{tables/ablations}

\section{Results}
\label{sec:results}

\noindent\textbf{Per-script performance:}
Table~\ref{tab:scriptwise} breaks down OCR and glyph count accuracy by script and region.
\paperName\ performs consistently across geographically and structurally diverse scripts: it attains low CER on  Newar (5.1),  Sharada (5.2), Kannada (5.5), Gurmukhi (5.5) and Malayalam (5.7), while remaining competitive on more visually complex scripts such as Grantha (12.4) and Telugu (12.3).
Performance drops are concentrated in complex scripts with dense ligatures and lower training line counts (Modi: 18.6; Siddham: 18.1). Qualitative results can be viewed in Fig.~\ref{fig:qualitativeresults}. For most scripts, performance in our default (jointly trained) setting is better than training dedicated per-script models (c.f. CER, Mono CER in Table~\ref{tab:scriptwise}), demonstrating the benefit of our unified design.

\noindent\textbf{Comparison with baselines:}
Table~\ref{tab:baseline_comparison} reports the overall performance of \paperName\ against classical OCR engines and recent HTR models on our \scriptcount\ scripts.
\paperName\ achieves \textbf{6.9\%} CER, significantly outperforming LSTM based models - Tesseract (32.4\%) and Kraken (28.5\%), and Transformer based models including VLT (19.6\%), HTR-VT (14.8\%), HTR-ConvText (12.9\%), and TrOCR (14.7\%).

\begin{figure}
    \centering
    \includegraphics[width=0.8\textwidth]{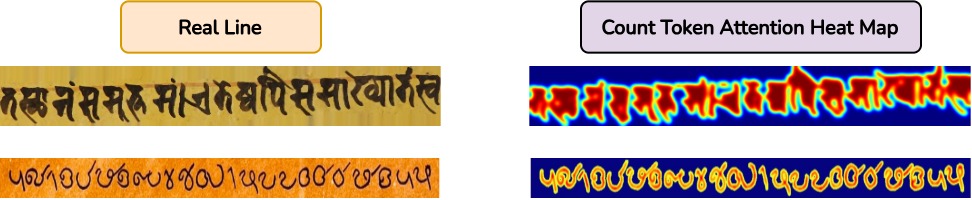}
    \caption{Count token embedding heat map}
    \label{fig:attentionmap}
    \vspace{-5pt}
\end{figure}

\noindent\textbf{Count-MAE for Glyph count prediction:}
In addition to transcription, \paperName\ predicts per line glyph count with an overall 1.1 Count-MAE.
Count errors track script difficulty: scripts with cleaner segmentation structure (e.g., Newar: 0.6, Sharada: 0.7) yield lower Count-MAE, while ligature-dense scripts exhibit higher error (e.g., Modi: 2.3). The heatmap visualization of attention structure induced from the count token in \paperName\ illustrates its utility in providing glyph-level information (see Fig.~\ref{fig:attentionmap}). The count can also act as a lightweight quality control cue for  evaluation, flagging unusual length predictions for analysis.

\noindent\textbf{F1 score for script classification:}
Script recognition is highly accurate in our experiments (macro-F1 close to 0.99), showing \paperName\ consistently identifies the script. This gives us reliable script metadata.

\noindent\textbf{Transferability:}
Table~\ref{tab:transfer_eval} evaluates how well \paperName\ serves as a pretrained initializer beyond the \scriptcount\ in-domain scripts.
First, fine tuning on out of domain benchmarks yields strong transfer: \paperName\ attains 6.9 , 7.2 , 8.1  CER on IAM, RIMES, LAM , 1.3 on Tibetan, 0.4 on Chinese, and 9.5 on the contemporary Indian handwriting dataset PhD-Indic~\cite{phd11}, suggesting that the synthetic pretraining learns reusable stroke ,  sequence level representations rather than script-specific templates.
Second, the unseen-script experiment (Table~\ref{tab:transfer_eval}b) shows that even a smaller \paperName-10 initialization performs competitively, indicating that our manuscript-realistic synthetic corpus provides a strong starting point that can be adapted with limited real data.
Increasing script diversity further improves this initialization and consistently lowers CER when fine tuning on new scripts (e.g., Sharada: 5.5$\rightarrow$5.2; Grantha: 12.6$\rightarrow$12.4).

\paperName\ achieves low average CER (6.9\%  - Tab.~\ref{tab:scriptwise}) while improving worst case robustness on low-resource scripts. In addition, it provides reliable auxiliary structural supervision through per line glyph counting (1.1 Count-MAE) and transfers effectively to non Indic writing systems. Overall, these results indicate that unified multi-script modeling is a scalable alternative to maintaining separate, script-specific OCR pipelines.

\section{Ablation Study}
\label{sec:ablations}

We conduct a systematic ablation study to quantify the contribution of each architectural and design component in \paperName. Unless otherwise specified, all variants are trained under identical settings.

\noindent \textbf{Output Representation:} We compare three output representations: Roman \texttt{WX} (default), IAST, and script-specific vocabularies. Using IAST results in 11.0 CER and 1.7 Count-MAE, while script-specific vocabularies degrade performance further to 14.1 CER and 1.9 Count-MAE due to limited cross-script information sharing. In contrast, Roman \texttt{WX} as achieves the best performance (6.9 CER, 1.1 Count-MAE), highlighting the benefit of a shared vocabulary.

\noindent \textbf{Auxiliary Tokens:} \paperName\ incorporates script identity and glyph-count tokens as auxiliary supervision to support training. Removing these signals increases CER to 9.7. The script token encourages the encoder to learn script aware representations, while the count token provides a global notion of text length that helps regulate errors. Together, these signals improve robustness on visually diverse historical manuscripts and also serve as useful indicators during annotation and validation.

\noindent \textbf{Encoder Depth:} We vary the Transformer encoder depth to study modeling capacity. A 4-layer encoder results in 9.3 CER and 1.5 Count-MAE, while increasing depth to 8 layers improves performance to 8.1 CER and 1.3 Count-MAE. However, our chosen configuration (6 layers) achieves the best balance with 6.9 CER and 1.1 Count-MAE.

\noindent \textbf{Text recognition loss} ($\mathcal{L_{OCR}}$):  Finally, we compare CTC loss (default) with cross-entropy (CE) loss. Replacing CTC with CE increases CER to 8.3 and Count-MAE to 1.6. Unlike CE, CTC does not require strict token-to-position alignment and naturally handles variable-length outputs.

\section{Conclusion}
We presented \paperName, the first  unified multi-script OCR framework for  Indic scripts. By combining manuscript-realistic synthetic pretraining, a unified Roman \texttt{WX} decoding target, and multitask supervision for script label and character count, the proposed model reduces the need for script-specific OCR pipelines.
Extensive experiments show competitive per-script accuracy across \scriptcount\ scripts, improved robustness for diverse manuscript lines, and promising cross-domain transfer beyond Indic scripts.

\bibliographystyle{splncs04}

\bibliography{references}

\end{document}

%% file: tables/unilipi15.tex
\begin{table}[t]
\centering
\scriptsize
\setlength{\tabcolsep}{5pt}

\caption{Script-wise performance of \paperName\ on \scriptcount\ scripts, grouped by geographic region.
We report \textbf{CER (\%)} for OCR transcription, \textbf{Mono CER (\%)} for a script-specific model trained only on that script (monolingual setting), and \textbf{Count-MAE} for glyph count prediction.}
\begin{tabular}{llrrr}
\toprule
\textbf{Region} & \textbf{Script} & \textbf{CER (\%)} & \textbf{Mono CER (\%)} & \textbf{Count-MAE} \\
\midrule

\multirow{3}{*}{\textit{North and Central India}}
& Devanagari   & 8.4  & 12.3 & 1.2 \\
& Sharada      & 5.2  & 6.9  & 0.7 \\
& Modi         & 18.6 & 17.4 & 2.3 \\

\midrule
\multirow{2}{*}{\textit{Western India}}
& Gurmukhi     & 5.5  & 6.5  & 1.2 \\
& Jaini        & 6.1  & 7.9  & 1.4 \\

\midrule
\multirow{2}{*}{\textit{Eastern India}}
& Bengali      & 18.3 & 17.1 & 2.2 \\
& Odia         & 16.3 & 15.1 & 1.8 \\

\midrule
\multirow{4}{*}{\textit{Southern India}}
& Grantha      & 12.4 & 12.6 & 1.7 \\
& Telugu       & 12.3 & 12.9 & 1.6 \\
& Malayalam    & 5.7  & 7.4  & 0.7 \\
& Kannada      & 5.5  & 6.9  & 1.1 \\

\midrule
\multirow{2}{*}{\textit{Himalayan}}
& Newar        & 5.1  & 7.1  & 0.6 \\
& Siddham      & 18.1 & 17.4 & 2.1 \\

\midrule
\textbf{Overall} & \textbf{All} & \textbf{6.9} & \textbf{8.6} & \textbf{1.1} \\
\bottomrule
\end{tabular}
\label{tab:scriptwise}
\end{table}

%% file: tables/baselines.tex
\begin{table}[t]
\centering
\scriptsize
\setlength{\tabcolsep}{5pt}
\renewcommand{\arraystretch}{1.15}

\caption{\textbf{Baseline comparison on UniLipi-\scriptcount.}
We evaluate \paperName\ (ours) against OCR baselines using overall \textbf{CER (\%)}.
To summarize script robustness, we additionally report the  lowest CER \textbf{(Script-L/CER-L)} and highest CER \textbf{(Script-H/CER-H)} scripts.
}
\begin{tabular}{l r c c r c c}
\toprule
\textbf{Model} &
\textbf{Params (M)} &
\textbf{CER} &
\textbf{Script-L} &
\textbf{CER-L} &
\textbf{Script-H} &
\textbf{CER-H} \\
\midrule

Kraken \cite{kraken2023}                  & 4.1   & 28.5 & Sharada & 15.2 & Bengali & 42.8 \\
PyLaia \cite{CRNN}                        & 5.3   & 24.3 & Newar   & 13.4 & Modi    & 38.7 \\
VLT \cite{Barrere2024VLT}                 & 7.0   & 19.6 & Malayalam & 10.8 & Bengali & 31.5 \\
Tesseract \cite{TesseractOCR}             & 8.0     & 32.4 & Malayalam & 18.6 & Modi    & 47.3 \\
HTR VT \cite{Li2025HTRVT}                 & 53.5  & 14.8 & Newar   & 7.9  & Modi    & 26.4 \\
HTR-ConvText \cite{truc2025htr}           & 65.9  & 12.9 & Sharada & 6.8  & Siddham & 23.7 \\
TrOCR \cite{Li2021TrOCR}                  & 385.0 & 14.7 & Newar   & 9.9  & Bengali & 23.8 \\

\midrule
\rowcolor{cyan!12}
\textbf{UniLipi (Ours)} & \textbf{85} & \textbf{6.9} & \textbf{Newar} & \textbf{5.1} & \textbf{Modi} & \textbf{18.6} \\

\bottomrule
\end{tabular}

\label{tab:baseline_comparison}
\end{table}

%% file: tables/t1.tex
\begin{table*}[t]
\centering
\tiny
\setlength{\tabcolsep}{4pt}
\renewcommand{\arraystretch}{1.1}

\caption{
\textbf{Transfer evaluation of \paperName-\scriptcount.}
(a) OCR across diverse writing systems; we report \textbf{CER (\%)} and \textbf{WER (\%)} after fine-tuning on each target benchmark.
(b) \paperName-10 is pretrained on 10 scripts (excluding Sharada, Jaini, and Grantha) and later fine-tuned on these unseen manuscript scripts, whereas \paperName-\scriptcount\ is pretrained on all \scriptcount\ scripts.
}
\label{tab:transfer_eval}

\begin{subtable}[t]{0.53\textwidth}
\centering
\begin{tabular}{l l r r}
\toprule
\textbf{Benchmark} & \textbf{Language} & \textbf{CER}  & \textbf{WER} \\
\midrule
Chi-Bench \cite{yu2021benchmarking}  & Chinese   & 0.4 & 2.1 \\
Tibetan Collections \cite{openpecha_ocr_2025}   & Tibetan   & 1.3  & 2.9\\
IAM \cite{IAM}              & English     & 6.9 & 24.5 \\ 
RIMES \cite{rimes_2011}             &  French     & 7.2 &  28.1 \\ 
LAM \cite{cascianelli2022lam}              & Italian   & 8.1 & 27.1 \\
READ \cite{toselli_2018_1297399} &  German & 8.6 & 26.4 \\
PhD-Indic \cite{phd11}        & Modern Indic     & 9.5    & 33.2  \\
Muharaf \cite{saeed2024muharaf} & Arabic & 10.5 & 35.1 \\
SleukRith \cite{Valy2018SleukRith} & Khmer & 15.3 & 44.1 \\
\bottomrule
\end{tabular}
\caption{Out-of-domain evaluation (only finetuning)}
\end{subtable}
\hfill
\begin{subtable}[t]{0.45\textwidth}
\centering
\begin{tabular}{l c c c}
\toprule
\textbf{Init.} & \textbf{Sharada} & \textbf{Jaini} & \textbf{Grantha} \\
\midrule
\paperName-10 & 5.5 & 6.1  & 12.6  \\
\rowcolor{cyan!12}
\textbf{\paperName-\scriptcount} &\textbf{ 5.2} &  \textbf{6.1}  &  \textbf{12.4} \\
\bottomrule
\end{tabular}
\caption{Unseen scripts in pretraining}
\end{subtable}

\end{table*}

%% file: tables/ablations.tex
\begin{table}[t]
\centering
\footnotesize
\setlength{\tabcolsep}{6pt}
\renewcommand{\arraystretch}{1.15}

\caption{
\textbf{Ablation study of UniLipi.}
Scores of final configuration is shown in the last row.
Each ablation modifies a single component while keeping all others identical.
}
\label{tab:ablations_compact}

\begin{tabular}{l l r c}
\toprule
\textbf{Modified Component} & \textbf{Default} & \textbf{CER} & \textbf{Count-MAE}  \\
\midrule

IAST as vocab &  Roman WX  & 11.0 & 1.7 \\
Dedicated vocab per script & Roman WX  & 14.1 & 1.9  \\

\midrule
No script token & Auxiliary tokens & 7.1 & 1.3 \\
No count token & Auxiliary tokens & 9.3 & - \\
No script and count token & Auxiliary tokens & 9.7 & - \\

\midrule
Encoder with 4 Layers & 6 Layers & 9.3 & 1.5 \\
Encoder with 8 Layers &  6 Layers & 8.1 & 1.3 \\

\midrule
$\mathcal{L_{OCR}}$ as Cross-Entropy &  CTC loss & 8.3 & 1.4 \\

\midrule
\rowcolor{cyan!12}
\textbf{UniLipi (Final)} & \textbf{Default design} & \textbf{6.9} & \textbf{1.1} \\

\bottomrule
\end{tabular}
\end{table}